\documentclass[letterpaper, 10 pt, conference]{ieeeconf}  

\IEEEoverridecommandlockouts                              

\usepackage{amsmath} 
\usepackage{amssymb}  
\usepackage{bm}
\usepackage{xcolor}
\usepackage{graphicx}
\usepackage{multirow}
\usepackage{booktabs}
\usepackage{makecell}
\usepackage{xcolor}
\usepackage{bbding}
\usepackage{subcaption}
\usepackage{hyperref}
\usepackage[linesnumbered,ruled,vlined]{algorithm2e}

\usepackage{xcolor}
\definecolor{lightblue}{RGB}{150,150,255}

\hypersetup{
    colorlinks=true,
    linkcolor=lightblue,
    citecolor=lightblue,
    urlcolor=lightblue
}

\SetKwBlock{Loop}{loop}{end loop}
\SetKwInput{KwInput}{Require}                

\SetCommentSty{mycommfont}
\def\algcomment#1{\textcolor[rgb]{0,0.6,0}{\# #1}}
\definecolor{darkgreen}{rgb}{0,0.4,0}

\title{\bf
Multi-Camera View Scaling for Data-Efficient Robot Imitation Learning
}

\author{Yichen Xie$^{1}$, Yixiao Wang$^{1}$, Shuqi Zhao$^{1}$, Cheng-En Wu$^{1}$, Masayoshi Tomizuka$^{1}$, Jianwen Xie$^{2}$, Hao-Shu Fang$^{3}$\\~\\$^1$UC Berkeley $^2$Lambda, Inc $^3$MIT}

\begin{document}

\maketitle


\begin{abstract}
The generalization ability of imitation learning policies for robotic manipulation is fundamentally constrained by the diversity of expert demonstrations, while collecting demonstrations across varied environments is costly and difficult in practice. In this paper, we propose a practical framework that exploits inherent scene diversity without additional human effort by scaling camera views during demonstration collection. Instead of acquiring more trajectories, multiple synchronized camera perspectives are used to generate pseudo-demonstrations from each expert trajectory, which enriches the training distribution and improves viewpoint invariance in visual representations. We analyze how different action spaces interact with view scaling and show that camera-space representations further enhance diversity. In addition, we introduce a multiview action aggregation method that allows single-view policies to benefit from multiple cameras during deployment. Extensive experiments in simulation and real-world manipulation tasks demonstrate significant gains in data efficiency and generalization compared to single-view baselines. Our results suggest that scaling camera views provides a practical and scalable solution for imitation learning, which requires minimal additional hardware setup and integrates seamlessly with existing imitation learning algorithms. The website of our project is \href{https://yichen928.github.io/robot_multiview/}{https://yichen928.github.io/robot\_multiview/}.
\end{abstract}    
\section{Introduction}
\label{sec:intro}
Recently, great advancements have been developed in training vision-based robot manipulation policy using imitation learning~\cite{osa2018algorithmic,chi2025diffusion,ze20243d} from human demonstrations. Unlike reinforcement learning, imitation learning does not allow policies to explore various environmental settings. Consequently, the generalization ability of imitation learning policy highly depends on the distribution of training data. If the training data are not diverse enough, the ability of policy learned from imitation learning will be strictly bounded.

To ensure the diversity of training data, there is an intuitive way that collects a large amount of human demonstrations. Although recent advances have been made in addressing this problem with improved hardware~\cite{chi2024universal,fang2025dexop} or cross-embodiment data utilization~\cite{khazatsky2024droid,o2024open,fang2023rh20t}, intensive human labor is still inevitable to scale up the size of the training data. As a result, it remains an open problem regarding how to maximize the diversity with limited data collection budget. Among these, one fundamental principle is emphasized in~\cite{hu2024data} that the generalization of the imitation learning policy is determined by \textit{the diversity of scenes} in the human demonstrations, which shows that the diversity of environments is far more important than the absolute number of demonstrations. However, in practice, arranging new environment and background layout is even more difficult and time-consuming in the data collection process than simply repeating the actions in the same scene. 


To this end, we seek for a more efficient way to improve the scene diversity with minimal efforts by making the full use of the limited human demonstrations.
We notice that the robot manipulation occurs in 3D space, but most existing expert demonstration datasets~\cite{o2024open,khazatsky2024droid} only provide a single-view camera (and a in-hand camera) as visual inputs, which is a waste of the inherent rich information contained by each expert demonstration.
As revealed in many works over the past decade~\cite{su2015multi,kanezaki2018rotationnet,huang2021spatio,zhang2021self}, different camera perspective views contain significant visual diversity even for the same scene. In this case, multiple camera view images can exploit the inherent diversity of existing scenes.

\begin{figure}[t]
    \centering
    \includegraphics[width=\linewidth]{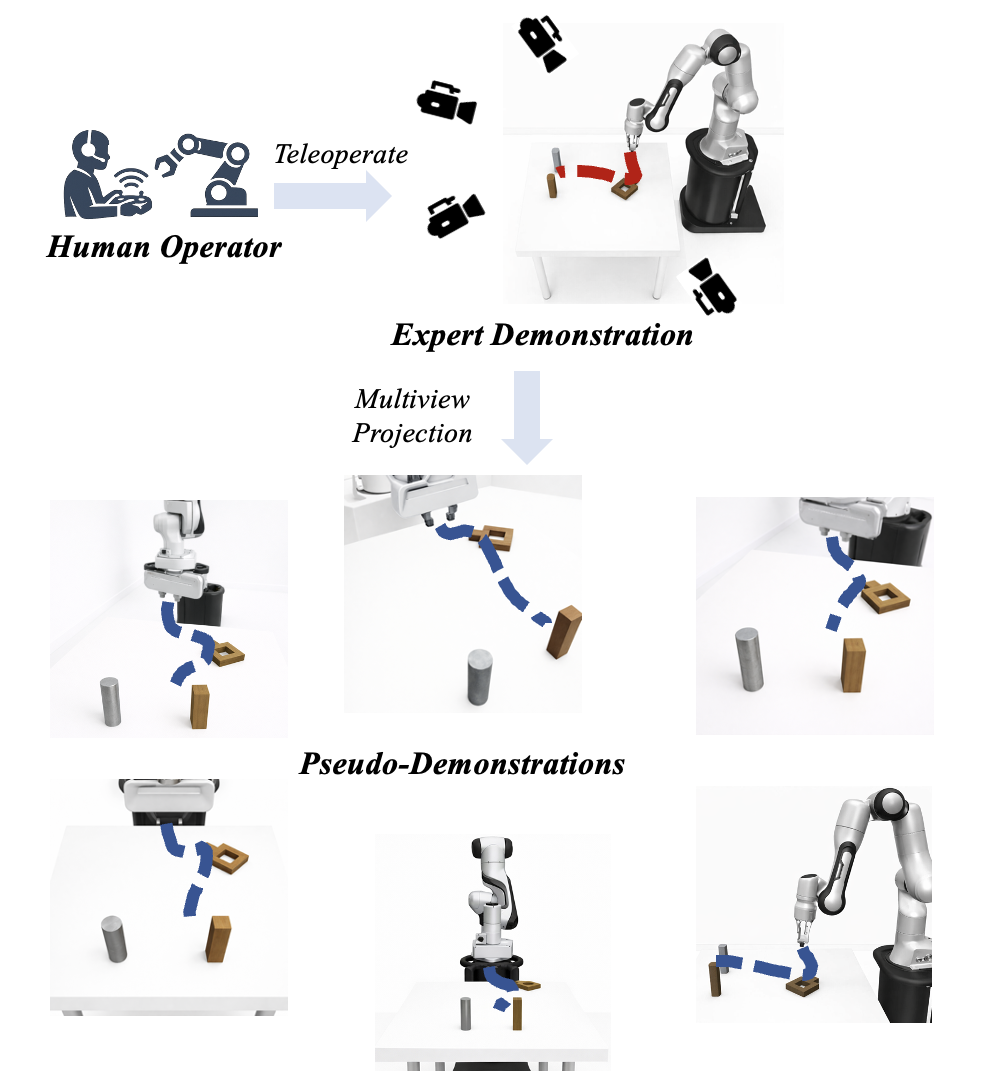}
    \vspace{-10pt}
    \caption{Camera View Scaling-Up. Each expert demonstration can be converted to multiple pseudo-demonstrations by exploiting multiple camera perspective views.}
    \label{fig:teaser}
    \vspace{-10pt}
\end{figure}

In this work, we propose that scaling up the camera views in the expert demonstration collection process (Fig.~\ref{fig:teaser}) has a similar effect to changing scenes on increasing data diversity and thus relieves the requirements for a large number of expert demonstrations in different environments. During training, we randomly select camera views for each scene to gather \textit{pseudo-demonstrations} from each expert demonstration. This design helps to reduce the intensive human labors in data collection since the number of teleoperation processes can decrease as long as multiple cameras are set up to capture the robot actions at the same time. We further take a deep dive into the effects of action spaces on the scaling up of camera views and show the great compatibility of our proposed training strategy to existing imitation learning algorithms. Our algorithm also produces a side product that the trained visual policy can benefit from multiple camera views in the inference stage through multiview action aggregation. We evaluate our algorithm in both simulation and real environment. Results show that our algorithm can greatly improve the data efficiency and reduce the human efforts in data collection in multiple situations. Our contributions can be summarized as follows:

\begin{itemize}
    \item We identify a missing point in existing human demonstrations that the inherent scene diversity is wasted in the single-view camera image. To this end, we propose a free-lunch strategy that scales up the camera views in the data collection process.
    \item We provide a deep insight into the effect of policy action space on scaling up camera views and further develop an aggregation algorithm to combine multiview visual information using a single-view policy during inference.
    \item We conduct extensive experiments in both simulation and real environment to evaluate our proposed algorithm. Results show the great effectiveness and compatibility of this algorithm.
\end{itemize}


\section{Related Work}
\label{sec:related_work}

\subsection{Imitation Learning for Robot Manipulation.}
The goal of imitation learning is to enable robots to learn complex skill from human expert demonstrations. This algorithm demonstrates promising performance in the field of robot manipulation~\cite{levine2016end,zitkovich2023rt,team2024octo,zhao2025dexh2r,chi2025diffusion}. Typically, most imitation learning algorithms include a perception module to extract useful information from observations and a policy module to generate actions based on the perception outputs. Recent methods~\cite{shang2024theia,xia2025cage} integrate pretrained vision foundation models~\cite{caron2021emerging,kirillov2023segment} as the visual encoder to improve the generalization ability, which leverages the large-scale pretraining on cross-domain data. For policy module, diffusion-based methods~\cite{chi2025diffusion,zhang2025flowpolicy} excel in modeling the complex and diverse action distributions. Some prior works~\cite{chi2024universal} also notices the influence of action space on the generalization of imitation learning policy. Despite all these advancements, the performance of imitation learning still highly relies on the amount and quality of expert demonstrations~\cite{hu2024data}. Orthogonal to previous research, we propose to make full use of multiview information of each expert demonstration to improve the data efficiency of imitation learning policies.

\subsection{Multiview Visual Representation Learning.} Over the past decade, great attention has been paid to exploiting multiview visual information. It is intuitive that cameras from different perspective views capture diverse visual information for the same scene. This principle sets up the foundation of image-based 3D scene reconstruction tasks~\cite{mildenhall2021nerf,kerbl20233d}. This cross-view diversity also helps the neural network to learn strong representation from multiview images. Early works \cite{su20153d,su2015render,wang2019perspective} consider images from different perspective views as a strong data augmentation method that compensates for the missing or biases in the original dataset. This philosophy furthers inspires some self-supervised algorithms to learn generalizable visual representations through cross-view contrastive learning~\cite{huang2021spatio,zhang2021self} or mask modeling~\cite{weinzaepfel2022croco,weinzaepfel2023croco}. In robot manipulation, many existing methods~\cite{sermanet2018time,seo2023multi,akinola2020learning} take multiview camera images as model inputs for collect complementary visual information, but this requires complex sensor setup in the deployment. In contrast, we scale up the camera views during training to improve the training data diversity with limited expert demonstrations, while our model can work with a single-view camera in the inference stage.
\section{Methodology}
In this section, we explain our proposed method in detail. In Sec.~\ref{sec:preliminary}, we start from current preliminary formulation of visuomotor policy. Then, in Sec.~\ref{sec:scale}, we explain how to scale up camera views in the imitation learning training process, where the effect of action space is discussed in Sec.~\ref{sec:space}. Finally, in Sec.~\ref{sec:aggregation}, we show that our policy can exploit multiview inputs in the inference stage.
 
\subsection{Preliminary: Imitation Learning for Visuomotor Policy}
\label{sec:preliminary}
The visuomotor policy aims to predict the future robot actions based on some visual observations.
Typically, its model can be divided into a perception encoder $g_{\phi}$ that extracts useful information from visual observations and a policy backbone $h_{\theta}$ that outputs the future robot actions.
\begin{equation}
    \hat{\mathbf{a}}=h_{\theta}\circ g_{\phi}(\mathbf{o})
    \label{eq:}
\end{equation}
Here, we mainly consider diffusion policy backbone~\cite{chi2025diffusion}, which models the conditional distribution $p_{\Theta}(\mathbf{a}|\mathbf{o})=\epsilon_{\Theta}(\mathbf{o})$.

For imitation learning, we collect a training set $\mathcal{T}_{demo}$ consisting of expert demonstrations. In most cases, each expert trajectory $\bm{\tau}_i$ is split into a set of pair-wise observations and actions $(\mathbf{o}_{i,t},\mathbf{a}_{i,t})$ along the temporal dimension. The model parameters are updated by minimizing the difference between predicted actions and expert demonstrations, and the size and diversity of expert demonstrations are critical for the generalization ability of the learned policy.
\begin{equation}
\begin{aligned}
&\hat{\Theta} = \arg\min_{\Theta} \ \mathbf{E}_{\mathbf{o}_{i,t},\mathbf{a}_{i,t}}
\left(\text{Loss}(\hat{\mathbf{a}}_{i,t}, \mathbf{a}_{i,t})\right) \\
&\hat{\mathbf{a}}_{i,t} = h_{\theta}\circ g_{\phi}(\mathbf{o}_{i,t})
\end{aligned}
\end{equation}

\subsection{Camera View Scaling-up for Pseudo-Demonstrations}
\label{sec:scale}
In most existing datasets, each expert demonstration consists of a single third-view camera observation. Each trajectory $\bm{\tau}_i$ with horizon $L_i$ is composed of a sequence of images $\mathbf{o}_{i,t}$ and actions $\mathbf{a}_{i,t}$, where $t=1,2,\dots,L_i$. In this case, the only way to expand the training set is to collect more expert demonstrations.

In contrast, we propose a simple way to utilize multiview visual observations as additional pseudo-demonstrations. Several cameras are installed at different perspectives when each expert demonstration is being collected. With multiview observations, each demonstration $\bm{\tau}_i^{MV}$ can be expanded into multiple pseudo-demonstrations $\{\bm{\tau}_i^{v}\}_{v=1}^V$ where $V$ is number of cameras. Further, each pseudo demonstration $\bm{\tau}_i^{v}$ is composed of a sequence of images $\mathbf{o}_{i,t}^v$ and actions $\mathbf{a}_{i,t}^v$. Accordingly, the expanded training set can be denoted as $\mathcal{T}_{demo}^{MV}=\{(\mathbf{o}_{i,t}^v,\mathbf{a}_{i,t}^v)\}_{v,t}$. It is worth mentioning that scaling up camera views adds negligible extra effort in expert demonstration collection process since each expert demonstration still requires only a single execution.

\begin{figure}[t]
    \centering
    \includegraphics[width=0.8\linewidth]{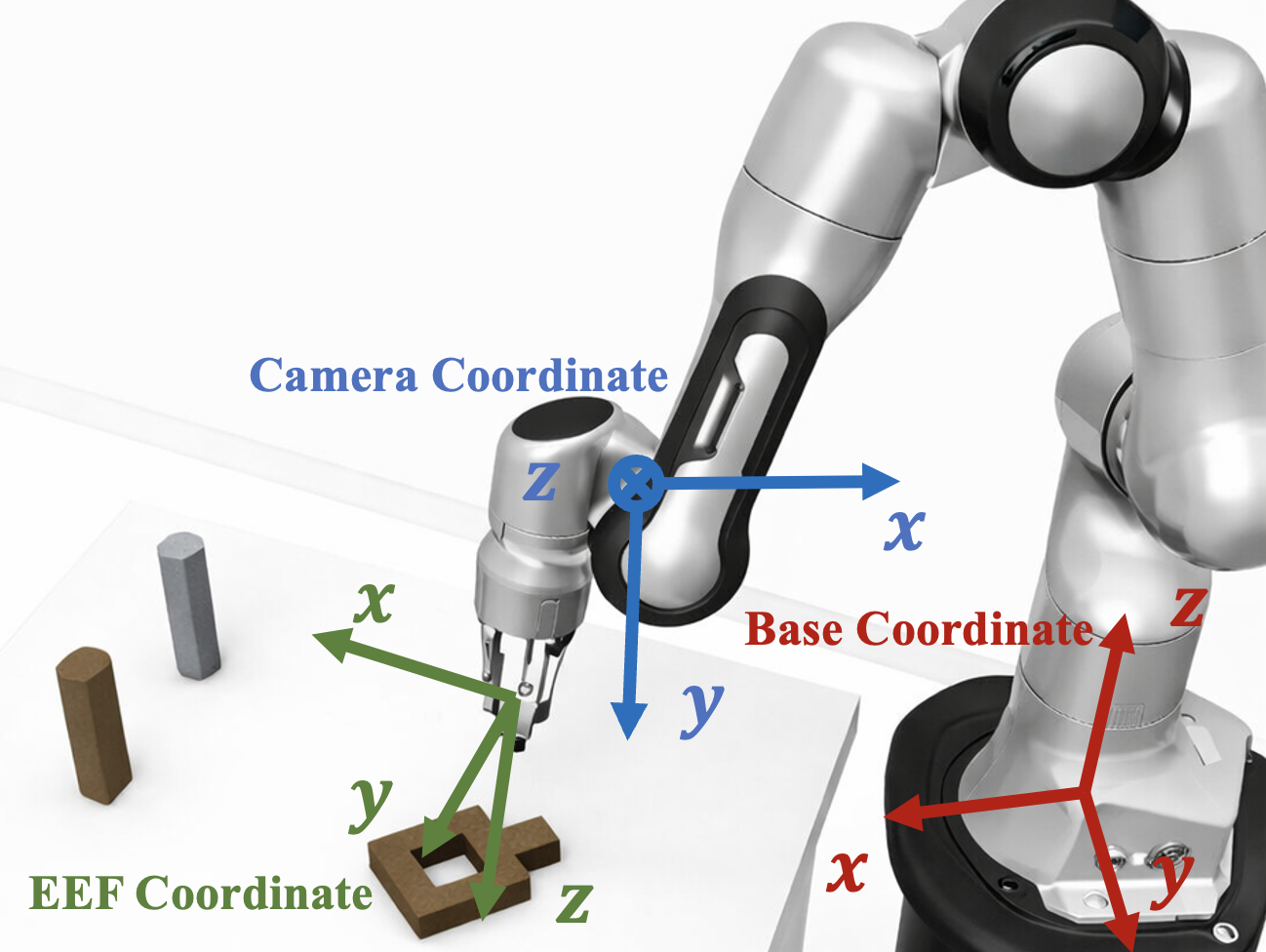}
    \caption{Action Space for Pseudo-Demonstrations. We consider three potential action spaces: \textit{base}, \textit{camera}, and \textit{EEF}.}
    \label{fig:space}
    \vspace{-10pt}
\end{figure}

\subsection{Robot Actions in View-Scaled Imitation Learning}
\label{sec:space}
The robot action is represented with the changes in end-effector poses $(\Delta\mathbf{p},\Delta\mathbf{R},s)$ at multiple future time steps~\cite{zhao2023learning} where $\Delta\mathbf{p}\in\mathbb{R}^3$ refers to the 3D translation, $\Delta\mathbf{R}\in SO(3)$ refers to the end-effector rotation, and $s\in\{0,1\}$ denotes the gripper state. To simplify the notion, $\mathbf{a}_{i,t}$ denotes the single-step action \textit{i.e.} $\mathbf{a}_{i,t}=(\Delta\mathbf{p}_{i,t},\Delta\mathbf{R}_{i,t},s_{i,t})$. By default, both location $\mathbf{p}$ and rotation $\mathbf{R}$ are represented in a fixed robot base coordinate.

When it comes to the derivation of pseudo-demonstration $\{\bm{\tau}_i^{v}\}_{v=1}^V$ from each expert demonstration, there are three potential solutions to determining robot action $\mathbf{a}_{i,t}^v$ depending on the action spaces (Fig.~\ref{fig:space}). We denote the transformation from original action $\mathbf{a}_{i,t}$ in the expert demonstration to pseudo-demonstration actions $\mathbf{a}_{i,t}^v$ as 
\begin{equation}
    \mathbf{a}_{i,t}^v=\mathcal{F}^v(\mathbf{a}_{i,t})
    \label{eq:transform}
\end{equation}
which is elaborated in different cases as below.

\paragraph{Robot Base Space} Regardless of camera perspectives in each pseudo-demonstration, all the actions share the same robot base coordinate fixed in the workspace, so the target action for each pseudo demonstration is a simple copy of the robot action in the original demonstration, \textit{i.e.,}
\begin{equation}
    \mathbf{a}_{i,t}^v=\mathbf{a}_{i,t},\forall v=\{1,2,\dots,V\}.
\end{equation}

\paragraph{End-Effector Space} As mentioned in~\cite{chi2024universal}, we can convert the robot action to the coordinate defined by the end-effector pose $(\mathbf{p}_{i,t},\mathbf{R}_{i,t})$ at each timestep $t$
\begin{equation}
    \Delta\mathbf{p}_{i,t}^{EEF}=\mathbf{R}_{i,t}^T\Delta\mathbf{p}_{i,t},\quad\Delta\mathbf{R}_{i,t}^{EEF}=\mathbf{R}_{i,t}^T\Delta\mathbf{R}_{i,t}\mathbf{R}_{i,t}.
\end{equation}
The action for each pseudo-demonstration is also unrelated to the camera perspectives as  
\begin{equation}
    \mathbf{a}_{i,t}^v=\mathbf{a}_{i,t}^{EEF},\forall v=\{1,2,\dots,V\}.
\end{equation}

\paragraph{Camera Space} To maximize the diversity of pseudo-demonstrations, we propose considering the robot actions in camera space. In this case, different pseudo demonstrations will have distinct actions even if they are derived from the same expert demonstration. For each pseudo-demonstration determined by camera view $v$, given the extrinsic parameters of each camera $(\mathbf{p}_v^{ext},\mathbf{R}^{ext}_v)$, we can convert the robot actions to the camera space as follows.
\begin{equation}
    \Delta\mathbf{p}_{i,t}^{v}=\mathbf{R}^{ext}_v\Delta\mathbf{p}_{i,t},\quad\Delta\mathbf{R}_{i,t}^{v}=\mathbf{R}^{ext}_v\Delta\mathbf{R}_{i,t}({\mathbf{R}_v^{ext}})^T
\end{equation}

\begin{algorithm}[tb!]
    \caption{\textbf{Multiview Policy Composition}}
    \label{alg:composition}
    \KwInput{Multiview observations $\{\mathbf{o}^v\}_{v=1}^V$, Diffusion policy model $\epsilon_{\Theta}$, Per-view action transformation $\mathcal{F}^v$}

    Initialize $\mathbf{a}^T_{aggr}$ from standard Gaussian noise\\
    \For{$t=T,T-1,\dots,1$}{
    Initialize denoising step $\delta^t\leftarrow0$\\
        \For{$v=1,2,\dots,V$}{
        $\delta^t\leftarrow \delta^t + \gamma\cdot\mathcal{F}^{v^{-1}}\left(\epsilon_{\Theta}(\mathbf{o}^v)\right)$\\
        \algcomment{Aggregating robot action prediction conditioned on each view.}
        }
        Denoising step: $\mathbf{a}^{t-1}_{aggr}=\mathbf{a}^{t}_{aggr}-\mathcal{F}^v(\delta^t)$\\
    }
    \textbf{Return} robot action $\mathbf{a}^0_{aggr}$
\end{algorithm}

\subsection{Multi-View Action Aggregation in Deployment}
\label{sec:aggregation}
Our model is designed to take inputs from a single third-view camera to output the robot policy, aiming to benefit from camera view scaling-up during training without multi-camera deployment during inference. Besides, as a side product, our trained policy can also exploit multiview camera images during inference through policy composition. d

Given several actions predicted by the same policy network conditioned on different camera views, we can exploit the complementary multiview information by combining their predicted action distributions in this intuitive manner.
\begin{equation}
    p_{\Theta}(\mathbf{a}_{aggr}|\{\mathbf{o}\}_{v=1}^V)\propto\prod_{v=1}^Vp_{\Theta}(\mathbf{a}|\mathbf{o}_v)
    \label{eq:composition}
\end{equation}
In this case, the final robot policy $p_{\Theta}(\mathbf{a}|\{\mathbf{o}\}_{v=1}^V)$ will prefer robot actions that have a high likelihood under the distributions conditioned on all the camera views, \textit{i.e.}, the robot actions predicted by the policy network with all the different views of camera inputs. This preference helps to exploit complementary multiview information to reduce uncertainty inside a single-view visuomotor policy.

Inspired by~\cite{wang2024poco}, this multiview policy composition in Eq.~\ref{eq:composition} can be implemented by extending the sampling procedure in the denoising process of diffusion policy models.
\begin{equation}
    \mathbf{a}^{t-1}_{aggr}=\mathbf{a}^t_{aggr}-\mathcal{F}^v\left(\gamma\cdot\sum_{v=1}^V\mathcal{F}^{v^{-1}}\left(\epsilon_{\Theta}(\mathbf{\mathbf{o}^v})\right)\right)
\end{equation}
where $\mathcal{F}^v(\cdot)$ is the action transformation defined in Eq.~\ref{eq:transform} and $\gamma$ is a hyperparameter. $\epsilon_{\Theta}$ is the trained diffusion policy model that takes inputs from a single-view camera. We also summarize this process in Alg.~\ref{alg:composition}.
\section{Experiments}

In this section, we aim to answer the following questions regarding the scaling-up of camera views.
\begin{itemize}
    \item[Q1] \textit{Can multiple pseudo-demonstrations improve imitation learning performance?}
    \item[Q2:] \textit{What is the effect of action space on camera view scaling-up?}
    \item[Q3:] \textit{Does the selection of camera views impact the scaling-up performance?}
    \item[Q4:] \textit{Can our policy trained with pseudo-demonstrations generalize to different camera views during inference?}
    \item[Q5:] \textit{Can we exploit multiview visual inputs in the inference stage with single-view policy?}
    \item[Q6:] \textit{Is this framework effective in real-world setting?}
\end{itemize}
We start from experiment setups in Sec.~\ref{sec:setup}, and then provide results and comparison in Sec.~\ref{sec:results}.

\subsection{Experiment Setup}
\label{sec:setup}
\paragraph{Tasks} We evaluate our proposed framework in simulation and real environments separately. For simulation environments, we select three tasks (\textit{square}, \textit{can}, \textit{lift}) from robomimic~\cite{robomimic2021}.  For real-world environment, we conduct experiments in \textit{water pouring} task. All the tasks are evaluated with success rate as metric.

\paragraph{Expert Demonstrations} To evaluate the performance of data scaling, we conduct several groups of experiments for each task with different numbers of expert demonstrations $N$ and camera views $V$, which lead to $N\cdot V$ pseudo-demonstrations. For simulation experiments, we randomly select demonstrations from the existing trajectories of robomimic~\cite{robomimic2021}. For real-world experiments, we collect expert demonstrations through teleoperation by ourselves.

\paragraph{Camera Views} Unless otherwise specified, we consider a camera with its perspective view directed to the front of the robot arm, denoted as $Cam_F$. We rotate the camera around the workspace center by $15^{\circ}$ towards the left, right, upper, and down sides separately, deriving four additional camera views denoted as $Cam_{FL},Cam_{FR},Cam_{FU},Cam_{FD}$ respectively. All these cameras will record the expert demonstration for imitation learning policy training. During inference, unless otherwise specified, we only adopt the front view $Cam_F$.

\paragraph{Baseline} We apply the diffusion policy~\cite{chi2025diffusion} as the default baseline. The model is trained separately on each task using single third-person view visual inputs. To improve data efficiency, we apply a pretrained DINOV3-base model~\cite{simeoni2025dinov3} as the visual encoder, which is finetuned with a LoRA module~\cite{hu2022lora} in imitation learning.


\begin{table*}
    \centering
    \caption{Camera View Scaling-Up Results in Simulation Environments.}
    \label{tab:sim_results}
    \resizebox{\linewidth}{!}{
    \begin{tabular}{c|c|ccc|ccc|c}
    \toprule
    \multirow{2}{*}{\textbf{Training View}} & \multirow{2}{*}{\textbf{Space}}  & \multicolumn{3}{c}{\textbf{Square}} & \multicolumn{3}{|c|}{\textbf{Can}} & \textbf{Lift} \\
    & & $N=10$ & $N=25$ &  $N=50$ & $N=10$ & $N=25$ &  $N=50$ & $N=10$\\
    \midrule
    $Cam_F$ & Base & 0.14 & 0.26 & 0.42 & 0.18 & 0.54 & 0.72 &  0.69 \\
    \midrule
    \multirow{2}{*}{$Cam_F,Cam_{FL},Cam_{FR}$} & Base & 0.16 (\textcolor{darkgreen}{14\%$\uparrow$}) & 0.31 (\textcolor{darkgreen}{19\%$\uparrow$}) & 0.47 (\textcolor{darkgreen}{12\%$\uparrow$}) & 0.27 (\textcolor{darkgreen}{50\%$\uparrow$}) & 0.63 (\textcolor{darkgreen}{17\%$\uparrow$}) & 0.78 (\textcolor{darkgreen}{8\%$\uparrow$}) & 0.79 (\textcolor{darkgreen}{14\%$\uparrow$}) \\
    & Camera & 0.17 (\textcolor{darkgreen}{21\%$\uparrow$}) & 0.29 (\textcolor{darkgreen}{12\%$\uparrow$})& 0.47 (\textcolor{darkgreen}{12\%$\uparrow$})  & 0.30 (\textcolor{darkgreen}{67\%$\uparrow$}) & 0.68 (\textcolor{darkgreen}{26\%$\uparrow$}) & 0.83 (\textcolor{darkgreen}{15\%$\uparrow$}) & 0.76 (\textcolor{darkgreen}{10\%$\uparrow$}) \\
    \midrule 
    \multirow{2}{*}{\makecell[c]{$Cam_F,Cam_{FL},Cam_{FR}$\\$Cam_{FU},Cam_{FD}$}} & Base & \textbf{0.18 (\textcolor{darkgreen}{29\%$\uparrow$})} & \textbf{0.42 (\textcolor{darkgreen}{62\%$\uparrow$})} & 0.55 (\textcolor{darkgreen}{31\%$\uparrow$}) & 0.32 (\textcolor{darkgreen}{78\%$\uparrow$}) & 0.65 (\textcolor{darkgreen}{20\%$\uparrow$}) & 0.80 (\textcolor{darkgreen}{11\%$\uparrow$}) & 0.80 (\textcolor{darkgreen}{16\%$\uparrow$})\\
    & Camera & \textbf{0.18 (\textcolor{darkgreen}{29\%$\uparrow$})} & 0.39 (\textcolor{darkgreen}{50\%$\uparrow$}) & \textbf{0.58 (\textcolor{darkgreen}{38\%$\uparrow$})} & \textbf{0.37 (\textcolor{darkgreen}{105\%$\uparrow$})} & \textbf{0.68 (\textcolor{darkgreen}{26\%$\uparrow$})} & \textbf{0.85 (\textcolor{darkgreen}{18\%$\uparrow$})} & \textbf{0.85 (\textcolor{darkgreen}{23\%$\uparrow$})}\\
    \bottomrule
    \end{tabular}
    }
\end{table*}

\begin{figure*}[t]
    \centering
    \begin{subfigure}{\linewidth}
        \centering
        \includegraphics[width=\linewidth]{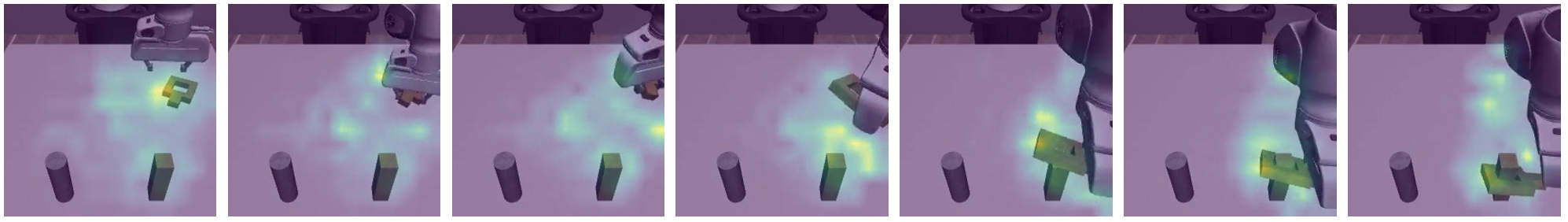}
        \caption{Training with 5 Camera Views}
    \end{subfigure}
    \begin{subfigure}{\linewidth}
        \centering
        \includegraphics[width=\linewidth]{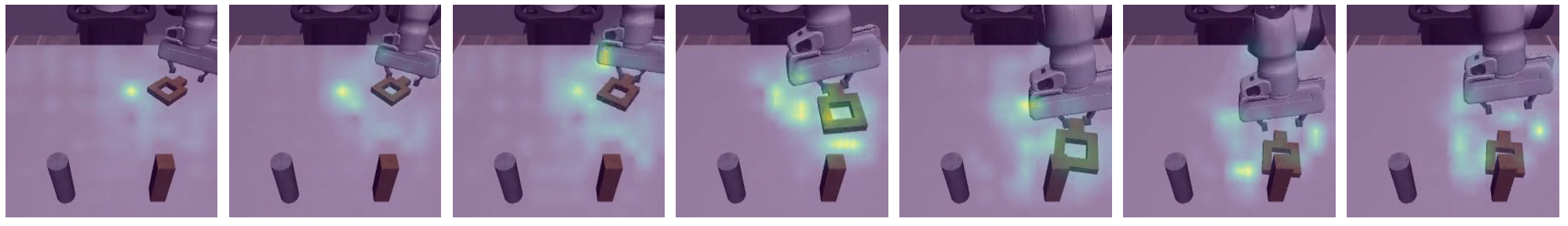}
        \caption{Training with Single Camera View}
    \end{subfigure}
    \caption{Visualization of Attention Scores. When trained with randomly picked camera views, the model learns to focus on the most important regions to finish the task. Otherwise, the attention is distracted to unrelated regions.}
    \label{fig:attn}
\end{figure*}

\subsection{Results}
\label{sec:results}
We conduct comprehensive experiments to answer the questions listed in the beginning of this section. The results are reported as follows.

\noindent\textit{\textbf{Q1: Can multiple pseudo-demonstrations improve imitation learning performance?}}

We report the results of three imitation learning tasks in Tab.~\ref{tab:sim_results}. The model is trained with different numbers of expert demonstrations. We compare the performance of policies trained with one ($Cam_F$), three ($Cam_F,Cam_{FL},Cam_{FR}$), and five ($Cam_F,Cam_{FL},Cam_{FR},Cam_{FU},Cam_{FD}$) camera views. It is clear that multi-camera training significantly outperforms single-camera training in all the scenarios, even achieving double success rates in some settings. Given the negligible human effort required to install extra cameras compared to collecting more expert trajectories, gathering pseudo-demonstrations from different perspective views serves as an economical and effective way to improve imitation learning performance. Additionally, training with five cameras achieves notably better performance than training with three cameras in most cases. This further reflects the benefits of scaling up camera views in imitation learning. 

To further justify the multi-camera training, we provide some qualitative analysis of transformer attention maps in Fig.~\ref{fig:attn}. It reveals that the model learns to pay more attention to important regions with multiple pseudo-demonstrations, which shows that the model can learn robust representation.

\begin{table}[tb]
    \centering
    \caption{Effect of Action Space.}
    \label{tab:action_space}
    \resizebox{\linewidth}{!}{
    \begin{tabular}{c|c|ccc}
    \toprule
       \multirow{2}{*}{\textbf{Training View}}  & \multirow{2}{*}{\textbf{Space}} & \multicolumn{3}{c}{\textbf{Square}}\\
       & & $N=10$ & $N=25$ & $N=50$  \\
       \midrule
        \multirow{2}{*}{$Cam_F$} & Base & \textbf{0.14} & \textbf{0.26} & \textbf{0.42}\\
        & EEF & 0.06 & 0.25 & 0.32 \\
        \midrule
       \multirow{2}{*}{\makecell[c]{$Cam_F,Cam_{FL},Cam_{FR}$\\$Cam_{FU},Cam_{FD}$}} & Base & \textbf{0.18} & \textbf{0.42} & \textbf{0.55}\\
       & EEF & 0.09 & 0.27 & 0.41 \\
    \bottomrule
    \end{tabular}
    }
\end{table}

\noindent\textit{\textbf{Q2: What is the effect of action space on camera view scaling-up?}}

In Tab.~\ref{tab:sim_results} and \ref{tab:action_space}, we consider three potential action spaces mentioned in Sec.~\ref{sec:space}, namely \textit{base}, \textit{camera}, and \textit{EEF}. Obviously, we find that the \textit{EEF} action space performs significantly worse than the other two counterparts no matter whether multiple camera views are adopted during training. This may be attributed to the moving origin of the EEF coordinate, which largely increases the difficulty of policy learning. For \textit{base} and \textit{camera} action spaces, they are inherently equivalent to each other when only a single fixed camera is utilized in training process. When it comes to multi-camera data collection and training with pseudo-demonstrations, notable performance gains are witnessed in both \textit{base} and \textit{camera} action spaces compared with single-view training. This fact reveals the robustness of scaling-up camera view with regard to action spaces. In addition, we also notice that \textit{camera} action space can bring better performance in most cases compared to \textit{base} action space. Under \textit{camera} action space, the projection to camera space allows multiple pseudo-demonstrations derived from the same expert demonstration to involve different robot actions as well as different visual inputs, thereby helping to improve the diversity of the training dataset composed of pseudo-demonstrations. However, it is also worth mentioning that \textit{camera} action space requires multiview camera calibration, which incorporates extra human labor. Therefore, it is a trade-off between performance and labor intensiveness to select the action spaces between \textit{base} and \textit{camera} spaces.


\begin{table}[tb]
    \centering
    \caption{Selection of Camera Views.}
    \label{tab:select_view}
    \vspace{-5pt}
    \begin{tabular}{c|ccc}
    \toprule
       \multirow{2}{*}{\textbf{Training View}} & \multicolumn{3}{c}{\textbf{Square}}\\
       & $N=10$ & $N=25$ & $N=50$  \\
       \midrule
        $Cam_F$ & 0.14 & 0.26 & 0.42\\
       \midrule
       $Cam_F,Cam_{FL}^{30},Cam_{FR}^{30}$ & \textbf{0.18} & 0.26  & 0.44 \\
       \midrule
       $Cam_F,Cam_T,Cam_S$ & 0.17 & 0.24 & 0.40 \\
        \midrule
       $Cam_F,Cam_{FL},Cam_{FR}$ & 0.17 & \textbf{0.29} & \textbf{0.47}\\
    \bottomrule
    \end{tabular}
\end{table}

\noindent\textit{\textbf{Q3: Does the selection of camera views impact the scaling-up performance?}}

In our default setting, multiple cameras are relatively close to each other. In this part, we consider two alternative multi-camera setups. 1) We apply three cameras with completely different perspective views: one in the front ($Cam_F$), one on the side ($Cam_S$), and one in the top ($Cam_T$). 2) We increase the relative angles between cameras from $15^{\circ}$ to $30^{\circ}$ denoted as $Cam_F,Cam_{FL}^{30},Cam_{FR}^{30}$.

Experiments results in these two settings are shown in Tab.~\ref{tab:select_view}. In setting 1), exploiting pseudo-demonstrations from multiple views hardly improves the performance of imitation learning. Since only $Cam_F$ is used during inference, the additional side-view and top-view cameras used during training capture perspectives that have distinct distributions. Despite their diversity, these out-of-distribution data may not help to improve the model performance with $Cam_F$ inputs during inference. In setting 2), a performance gain is also observed, but it is less significant than that in our default setting. We attribute this to a similar reason as in setting 1) that the additional camera views capture visual inputs that have a different distribution with visual inputs from $Cam_F$ during inference, so it cannot greatly improve the performance of a fixed view during inference. 

\begin{table}[tb!]
    \centering
    \caption{Camera View Generalization in the Inference.}
    \label{tab:multiview_infer}
    \begin{tabular}{c|c|ccc}
    \toprule
    \multirow{2}{*}{\textbf{Training View}} & \multirow{2}{*}{\textbf{Inference View}} & \multicolumn{2}{c}{\textbf{Can}}\\
    & &  $N=10$ & $N=25$ \\
    \midrule
      \multirow{3}{*}{\textit{Five Views}}& $Cam_F$  &  \textbf{0.37} & 0.68 \\
      & $Cam_{FL}$ & 0.30 & 0.66\\
      & $Cam_{FU}$ & \textbf{0.37} & \textbf{0.77}  \\
       \midrule
       \multirow{3}{*}{\textit{Single View}} & $Cam_F$ & 0.18 & 0.54\\
       &  $Cam_{FL}$ & 0.00 & 0.07  \\
       &  $Cam_{FU}$ & 0.01 & 0.05\\
    \bottomrule
    \end{tabular}
\end{table}

\begin{table*}
    \centering
    \caption{Multiview Composition with Single-View Policy.}
    \label{tab:multiview}
    \vspace{-5pt}
    \begin{tabular}{c|c|ccc|ccc}
    \toprule
    \multirow{2}{*}{\textbf{Training View}} & \multirow{2}{*}{\textbf{Inference View}}  & \multicolumn{3}{c}{\textbf{Square}} & \multicolumn{3}{|c}{\textbf{Can}} \\
    & & $N=10$ & $N=25$ &  $N=50$ & $N=10$ & $N=25$ &  $N=50$ \\
    \midrule
    $Cam_F$ & $Cam_F$ & 0.14 & 0.26 & 0.42 & 0.18 & 0.54 & 0.72 \\
    \midrule
    \multirow{2}{*}{$Cam_F,Cam_{FL},Cam_{FR}$} & $Cam_F$ & 0.17  & 0.29 & 0.47  & 0.30 & 0.68  & 0.83\\
    & $Cam_F,Cam_{FL},Cam_{FR}$ & \textbf{0.18} & \textbf{0.33} & \textbf{0.51} & \textbf{0.39} & \textbf{0.73} & \textbf{0.85}\\
    \bottomrule
    \end{tabular}
    \vspace{-15pt}
\end{table*}

\noindent\textit{\textbf{Q4: Can our policy trained with pseudo-demonstrations generalize to different camera views during inference?}}

Thanks to the pseudo-demonstrations from different camera views, our imitation learning policy can naturally generalize to different camera perspective views in the inference stage. As shown in Tab.~\ref{tab:multiview_infer}, our policy achieves stable performance over several different camera view inputs, which allows more flexibility in the camera setup in the deployment. In contrast, the policy trained only with a fixed view ($Cam_F$) does not have a reasonable generalization ability to other views in the deployment.

\noindent\textit{\textbf{Q5: Can we exploit multiview visual inputs in the inference stage with a single-view policy?}}

Although our framework is targeted for a single-view policy, as mentioned in Sec.~\ref{sec:aggregation}, it can exploit the aggregation of multiview camera inputs in the inference stage. Our single-view policy can take several different views as inputs and predict the robot actions independently in parallel, then the multiple robot actions are aggregated together following Alg.~\ref{alg:composition}. The results are reported in Tab.~\ref{tab:multiview}. We find that using three views in the deployment can improve the model performance compared with using a single fixed view. Since these multiview inputs are processed independently, it enables high parallelism for the aggregation algorithm on GPUs, which reduces the extra latency introduced by the multiview aggregation. We highlight that the main focus of our algorithm is to train a single-view policy, and this multiview aggregation is considered as a free-lunch performance gain when multiview inputs are available during inference.

\begin{figure}[t]
    \centering
    \includegraphics[width=0.9\linewidth]{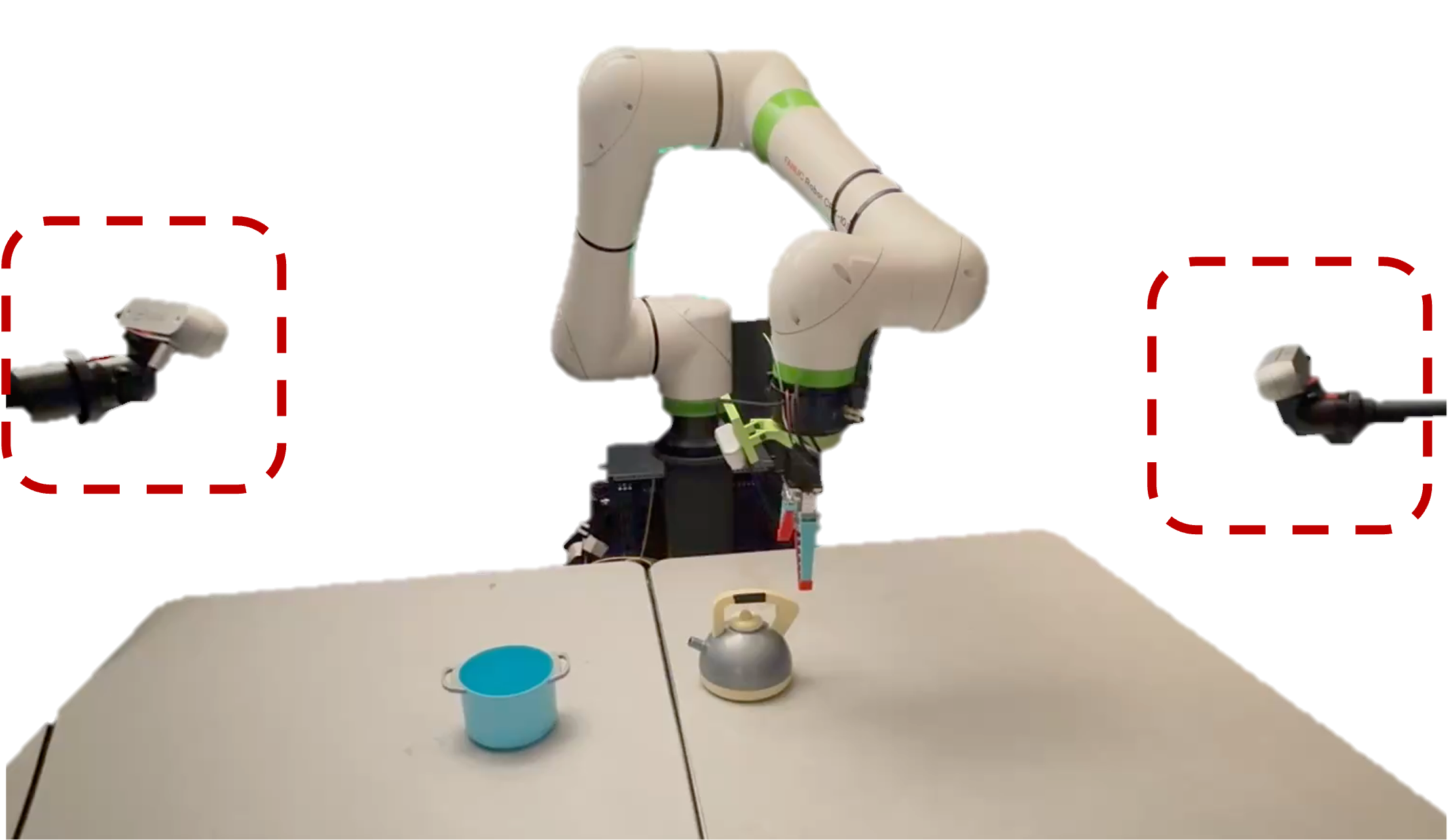}
    \vspace{-5pt}
    \caption{Real-World Robot Setup. We use a FANUC CRX-10iA robotic arm equipped with a parallel-jaw gripper with silicone soft fingertips and two camera views.}
    \label{fig:real_world_setup}
\end{figure}

\begin{table}[tb!]
    \centering
    \caption{Real-World Experiment Results. We consider whether to train with pseudo-demonstration and infer with multiview aggregation.}
    \label{tab:real}
    \begin{tabular}{c|c|cc}
    \toprule
       \textbf{Training View}  & \textbf{Composition} & $N=25$ & $N=50$  \\
       \midrule
        \textit{Single} & \XSolidBrush & 0.45 & 0.70 \\
        \midrule
        \multirow{2}{*}{\textit{Both}} & \XSolidBrush & 0.50 & \textbf{0.85} \\
         & \Checkmark & \textbf{0.75} & \textbf{0.85}\\
    \bottomrule
    \end{tabular}
    \vspace{-5pt}
\end{table}

\noindent\textit{\textbf{Q6: Is this framework effective in real-world setting?}}

\paragraph{Real-World Experiment Setup} We conduct real-robot experiments using a FANUC CRX-10iA robotic arm equipped with a parallel-jaw gripper with silicone soft fingertips. The task requires grasping a teapot and pouring its contents into a cooking pot. At the start of each trial, both the teapot and the pot are randomly positioned and oriented within predefined workspace bounds. Observations consist of RGB images from two fixed cameras (left-view and right-view). The action space comprises end-effector Cartesian motion (translation and orientation) together with a gripper open/close command. We collect 50 expert demonstrations through teleoperation and train policies with either 25 or 50 demonstrations. We train a diffusion policy model~\cite{chi2025diffusion} with ResNet-18~\cite{he2016deep} and Spatial Softmax~\cite{levine2016end} for the visual encoder and FiLM~\cite{perez2018film} for visual conditioning.  

\paragraph{Experiment results} We compare the single-view policy trained with data collected from a single view or both views. Results in Tab.~\ref{tab:real} show that training with both views jointly yields better performance. However, due to hardware limitations, we only have two camera views in total. As a result, the performance gain from two-view training is not very significant. Afterwards, we further study whether multiview composition in the inference stage can better improve the visual policy performance. The results are also revealed in Tab.~\ref{tab:real}. Multiview composition can help improve model performance over single-view inference, which verifies the effectiveness of our algorithm in Alg.~\ref{alg:composition} in the real-world setting, suggesting that we can exploit complementary viewpoint-specific cues.

\section{Conclusion}

In this work, we introduced camera view scaling-up as a simple yet effective way to improve data efficiency in imitation learning for robot manipulation. Instead of collecting additional expert demonstrations, we leverage synchronized multiview observations to generate pseudo-demonstrations, thereby increasing inherent scene diversity with negligible additional human effort. We systematically analyzed how action space interacts with view scaling and showed that both base and camera  action spaces benefit from multiview training, while camera-space actions further enhance diversity. Moreover, we proposed a multiview action aggregation mechanism that enables a single-view policy to exploit multiple cameras at inference time without architectural modification. Extensive experiments in simulation and real-world environments demonstrate consistent improvements in success rates across tasks and data regimes. Our results suggest that scaling camera views, rather than scaling demonstrations, offers a practical and scalable pathway toward more data-efficient visuomotor policy learning. While our framework focuses on viewpoint diversity within a fixed scene, future work may explore combining view scaling with environment randomization, cross-embodiment data, or large-scale foundation models to further enhance generalization. We hope this work encourages rethinking data scaling strategies in imitation learning and highlights the untapped potential of multiview supervision in robot policy learning.

\bibliographystyle{IEEEtran}
\bibliography{IEEEabrv}

\end{document}